%% file: main.tex
\documentclass[sigconf]{acmart}

\usepackage{amsmath,amsfonts}
\usepackage[flushmargin]{footmisc}
\usepackage{tikz}
\usepackage{forest}
\usetikzlibrary{arrows,shapes,positioning,shadows,trees}

\usepackage[algo2e,linesnumbered,ruled,lined,noend]{algorithm2e}
\SetKwInOut{Parameter}{Parameters}
\SetKwInOut{Constraint}{Constraints}
\SetKwFunction{Mining}{Mining}
\SetKwFunction{MiningTwo}{Mining2}
\SetKwFunction{Candidates}{Candidates}
\SetKwProg{Fn}{Function}{}{}
\usepackage{graphicx}
\usepackage{textcomp}
\usepackage{color}
\usepackage{xcolor}
\usepackage{xspace}
\usepackage{subfigure}
\usepackage{enumerate}
\usepackage{pifont}
\def\BibTeX{{\rm B\kern-.05em{\sc i\kern-.025em b}\kern-.08em
    T\kern-.1667em\lower.7ex\hbox{E}\kern-.125emX}}
    
\usepackage{multirow} 

\usepackage[normalem]{ulem}
\useunder{\uline}{\ul}{}

\usepackage{pgf}
\usepackage{enumitem}
\setlist[itemize]{leftmargin=*}
\usepackage{stfloats}
\usepackage{array}
\usepackage{tabularx,booktabs}
\newcolumntype{P}[1]{>{\centering\arraybackslash}p{#1}}

\usepackage{makecell}

\usepackage{bbold}
\usepackage{mathtools}
\usepackage{comment}
\usepackage{microtype}
\usepackage[misc]{ifsym}
\usepackage{url}

\SetCommentSty{mycommfont}

\usepackage{contour}
\usepackage[normalem]{ulem}
\contourlength{0.8pt}
\newcommand{\myuline}[1]{%
  \uline{\phantom{#1}}%
  \llap{\contour{white}{#1}}%
}
\newcommand{\smallsection}[1]{{\noindent {\textbf{\myuline{#1}}}}}

\newcommand{\yes}{\textcolor{cyan}{\ding{52}}}

\input{dfn}

\AtBeginDocument{%
  \providecommand\BibTeX{{%
    \normalfont B\kern-0.5em{\scshape i\kern-0.25em b}\kern-0.8em\TeX}}}


\copyrightyear{2024}
\acmYear{2024}
\setcopyright{acmlicensed}\acmConference[KDD '24]{Proceedings of the 30th ACM SIGKDD Conference on Knowledge Discovery and Data Mining}{August 25--29, 2024}{Barcelona, Spain}
\acmBooktitle{Proceedings of the 30th ACM SIGKDD Conference on Knowledge Discovery and Data Mining (KDD '24), August 25--29, 2024, Barcelona, Spain}
\acmDOI{10.1145/3637528.3671457} 
\acmISBN{979-8-4007-0490-1/24/08} 

\begin{document}

    \title[A Survey on Hypergraph Neural Networks: An In-Depth and Step-By-Step Guide]{A Survey on Hypergraph Neural Networks: \\An In-Depth and Step-by-Step Guide}
    
	
	\author{Sunwoo Kim}
        \authornote{Equal contribution}
	\affiliation{%
    	\institution{KAIST}
            \city{Seoul}
            \country{Republic of Korea}
	}
	\email{kswoo97@kaist.ac.kr}

        \author{Soo Yong Lee}
        \authornotemark[1]
	\affiliation{%
		\institution{KAIST}
            \city{Seoul}
            \country{Republic of Korea}
	}
	\email{syleetolow@kaist.ac.kr}

        \author{Yue Gao}
	\affiliation{%
		\institution{Tsinghua University}
            \city{Beijing}
            \country{China}
	}
	\email{gaoyue@tsinghua.edu.cn}

        \author{Alessia Antelmi}
	\affiliation{%
		\institution{University of Turin}
            \city{Turin}
            \country{Italy}
	}
	\email{alessia.antelmi@unito.it}

        \author{Mirko Polato}
	\affiliation{%
		\institution{University of Turin}
            \city{Turin}
            \country{Italy}
	}
	\email{mirko.polato@unito.it}
	
	\author{Kijung Shin}
        \authornote{Corresponding author}
	\affiliation{%
		\institution{KAIST}
            \city{Seoul}
            \country{Republic of Korea}
	}
	\email{kijungs@kaist.ac.kr}

        \begin{abstract}
		\input{000Abstract}
	\end{abstract}
	
	\maketitle

    \vspace{10mm}
        
        \section{Introduction}
        \label{sec:introduction}
	\input{001intro}

	\section{Preliminaries}
        \label{sec:preliminaries}
        \input{002prelim}

	\section{Encoder Design Guidance}
        \label{sec:encoding}

\input{003encoding}
    
	\section{Objective Design Guidance}
        \label{sec:training}
        \input{004training}

        \section{Application Guidance}
        \label{sec:application}
        \input{005applications}

        \vspace{-1mm}
        \section{Discussions}
        \label{sec:discussions}
        \input{006discussions}

        \section*{Acknowledgements}
        {\small This work was partly supported by Institute of Information \& Communications Technology Planning \& Evaluation (IITP) grant funded by the Korea government (MSIT)  (No. 2022-0-00157, Robust, Fair, Extensible Data-Centric Continual Learning) (No. RS-2019-II190075, Artificial Intelligence Graduate School Program (KAIST)). This work has been partially supported by the spoke ``FutureHPC \& BigData” of the ICSC – Centro Nazionale di Ricerca in High-Performance Computing, Big Data and Quantum Computing funded by European Union – NextGenerationEU.
        }



        
        \bibliographystyle{ACM-Reference-Format}
        \balance
	\bibliography{000Ref}

\end{document}


\input{100config}

\title{\Large RASP: Robust Mining of Frequent Temporal Sequential Patterns under Temporal Variations - Supplementary Document}

\date{}

\maketitle

    \appendix
    \input{090appendix}
    

%% file: dfn.tex
\newcommand{\vb}{\boldsymbol{b}}

\newcommand{\vm}{\boldsymbol{m}}

\newcommand{\vp}{\boldsymbol{p}}
\newcommand{\vq}{\boldsymbol{q}}

\newcommand{\vw}{\boldsymbol{w}}
\newcommand{\vx}{\boldsymbol{x}}
\newcommand{\vy}{\boldsymbol{y}}

\newcommand{\mh}{\mathbf{H}}

\newcommand{\matp}{\mathbf{P}}
\newcommand{\mq}{\mathbf{Q}}

\newcommand{\mx}{\mathbf{X}}
\newcommand{\my}{\mathbf{Y}}

\newcommand{\calc}{\mathcal{C}}
\newcommand{\cald}{\mathcal{D}}
\newcommand{\cale}{\mathcal{E}}

\newcommand{\calg}{\mathcal{G}}

\newcommand{\caln}{\mathcal{N}}

\newcommand{\calv}{\mathcal{V}}

%% file: 000abstract.tex
Higher-order interactions (HOIs) are ubiquitous in real-world complex systems and applications. Investigation of deep learning for HOIs, thus, has become a valuable agenda for the data mining and machine learning communities. 
As networks of HOIs are expressed mathematically as hypergraphs, hypergraph neural networks (HNNs) have emerged as a powerful tool for representation learning on hypergraphs.
Given the emerging trend, we present the first survey dedicated to HNNs, with an in-depth and step-by-step guide.
Broadly, the present survey overviews HNN architectures, training strategies, and applications.
First, we break existing HNNs down into four design components:
(\textit{i}) input features, (\textit{ii}) input structures, (\textit{iii}) message-passing schemes, and (\textit{iv}) training strategies.
Second, we examine how HNNs address and learn HOIs with each of their components.
Third, we overview the recent applications of HNNs in recommendation, bioinformatics and medical science, time series analysis, and computer vision.
Lastly, we conclude with a discussion on limitations and future directions.

%% file: 001intro.tex
\begin{figure}[t]
    \centering
    \small
    \subfigure[Co-authors of publications]
    {\includegraphics[width=0.18\textwidth]{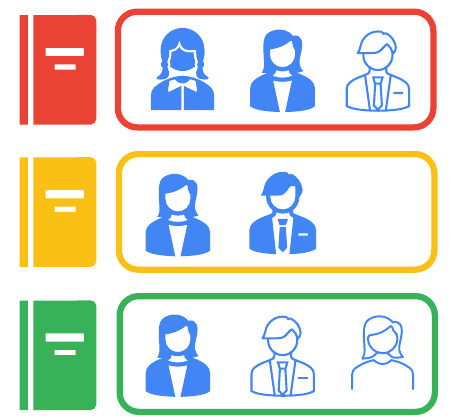}}
    \hspace{3mm}
    \subfigure[Hypergraph]
    {\includegraphics[width=0.18\textwidth]{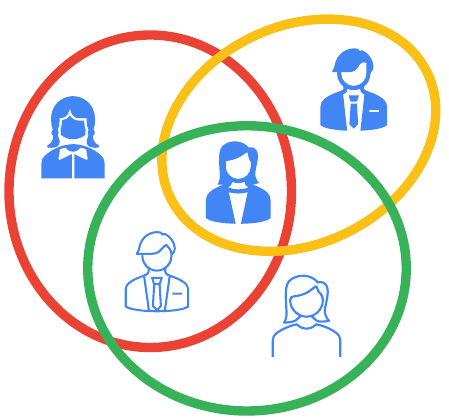}}
    \vspace{-2mm}
    \caption{\label{fig:examplehypergraphs}
    {An example hypergraph modeling the co-authorship relationship among five authors across three publications. 
    Each node represents an author, while each hyperedge includes all co-authors of a publication.}}
\end{figure}

Higher-order interactions (HOIs) are pervasive in real-world complex systems and applications. 
These relations describe multi-way or group-wise interactions, occurring from physical systems~\cite{battiston2022higher}, microbial communities~\cite{morin2022higher}, brain functions~\cite{expert2022higher}, and social networks~\cite{iacopini2022group}, to name a few. 
HOIs reveal {structural} patterns unobserved in their pairwise counterparts and inform network dynamics.
For example, they have been shown to affect or correlate with synchronization in physical systems~\cite{battiston2020networks}, bacteria invasion inhibition in microbial communities~\cite{mickalide2019higher}, cortical dynamics in brains~\cite{yu2011higher}, and contagion in social networks~\cite{de2020social}.

Hypergraphs mathematically express higher-order networks or networks of HOIs~\cite{bianconi2021higher}, 
where nodes and hyperedges respectively represent entities and their HOIs.
In contrast to an edge connecting only two nodes in pairwise graphs, a hyperedge can connect any number of nodes, offering hypergraphs advantages in their descriptive power.
For instance, as shown in Fig.~\ref{fig:examplehypergraphs}, the co-authorship relations among researchers can be represented as a hypergraph. 
With their expressiveness and flexibility, hypergraphs have been routinely used to model higher-order networks in various domains~\cite{de2020social, hao2024hypergraph, battiston2021physics, feng2021hypergraph} to uncover their structural patterns~\cite{lee2024survey, kim2023reciprocity, kim2023transitive, lee2020hypergraph, do2020structural, lee2021hyperedges}. 

As hypergraphs are extensively {utilized}, the demand grew to make predictions on them, estimating node properties or identifying missing hyperedges.
Hypergraph neural networks (HNNs) have shown strong promise in solving such problems.
For example, they have shown state-of-the-art performances in industrial and scientific applications, 
including missing metabolic reaction prediction~\cite{chen2023teasing},
brain classification~\cite{ji2022fc}, 
traffic forecast~\cite{zhao2023dynamic}, 
product recommendation~\cite{ji2020dual}, 
and more~\cite{han2023vision, xu2022counterfactual, ma2022learning}.

The research on HNNs has been exponentially growing.
Simultaneously, further research on deep learning for higher-order networks is an imminent agenda for the data mining and machine learning communities~\cite{papamarkou2024position}.
Therefore, we provide a timely survey on HNNs {that addresses} the following questions:
\begin{itemize}[leftmargin=*, noitemsep]
    \item \textbf{Encoding (Sec.~\ref{sec:encoding}).} 
    {\textit{How do HNNs effectively capture HOIs?}}
    \item \textbf{Training (Sec.~\ref{sec:training}).} 
    {\textit{How to encode HOIs with training objectives, especially when {external} labels are scarce or absent?}}
    \item \textbf{Application (Sec.~\ref{sec:application}).} \textit{What are notable applications of HNNs?}
\end{itemize}

\noindent Our scope is largely confined to HNNs for undirected, static, and homogeneous hypergraphs, with node classification or hyperedge prediction as their downstream tasks. 
The survey aims to provide an \textit{in-depth} and \textit{step-by-step} guide, with HNNs' design components (see Fig.~\ref{fig:overall_taxonomy}) and their analysis (see Table~\ref{tab:hnntaxonomy}).


%% file: 002prelim.tex
\input{000_taxonomy_tree}

In this section, we present definitions of basic concepts related to hypergraphs and HNNs.
See Table~\ref{tab:notations} for frequently-used symbols. 

A \textit{hypergraph} $\calg = (\calv, \cale)$ is defined as a set of nodes $\calv = \{v_{1},v_{2},\cdots , v_{\vert \calv \vert}\}$ and a set of hyperedges $\cale = \{e_{1}, e_{2},\cdots , e_{\vert \cale \vert}\}$.
Each hyperedge $e_{j}$ is a non-empty subset of nodes (i.e., $\emptyset \neq e_{j} \subseteq \calv$).
Alternatively, $\mathcal{E}$ can be represented with an incidence matrix $\mathbf{H} \in \{0,1\}^{\vert \calv \vert \times \vert \cale \vert}$, where $\mathbf{H}_{i,j} = 1$ if $v_{i} \in e_{j}$ and $0$ otherwise.
{The incident}  hyperedges of a node $v_{i}$, denoted as $\caln_{\cale}(v_{i})$, is the set of hyperedges that contain $v_{i}$ (i.e., $\caln_{\cale}(v_{i}) = \{e_{k} \in \cale : v_{i} \in e_{k}\}$).
We assume that each node $v_{i}$ and hyperedge $e_{j}$ are equipped with (input) node features $\vx_{i} \in \mathbb{R}^{d}$ and hyperedge features $\vy_{j} \in \mathbb{R}^{d'}$, respectively.\footnote{Sometimes, (external) node and hyperedge features may not be given. In such cases, one may utilize structural or identity features, as described in Sec.~\ref{subsec:features}.}
Similarly, we denote node and hyperedge feature matrices as $\mathbf{X} \in \mathbb{R}^{ \vert \calv\vert \times d}$ and $\mathbf{Y} \in \mathbb{R}^{ \vert \cale\vert \times d'}$, respectively, where the $i$-th row $\mathbf{X}_{i}$ corresponds to $\vx_{i}$ and $j$-th row $\mathbf{Y}_{j}$ corresponds to $\vy_{i}$.
In Sec.~\ref{subsec:features}, we detail approaches to obtain the features.

\input{000_table_notations}

\textit{Hypergraph neural networks} (HNNs) are neural functions that transform given nodes, hyperedges, and {their features} into vector representations {(i.e., embeddings)}.
Typically, their input is represented as either $(\mathbf{X}, \cale)$ or $(\mathbf{X}, \mathbf{Y}, \cale)$.
HNNs first prepare the input hypergraph structure $\cale$ (Sec.~\ref{subsec:express}).
Then, HNNs perform message passing between nodes (and/or hyperedges) to update their embeddings (Sec.~\ref{subsec:messagepassing}). 
A node (or hyperedge) message roughly refers to its vector representation for other nodes (or hyperedges) to aggregate.
The message passing operation is repeated $L$ times, where each iteration corresponds to one HNN layer. 
Here, we denote the $\ell$-th layer embedding matrix of nodes and hyperedges as $\mathbf{P}^{(\ell)} \in \mathbb{R}^{\vert \calv \vert \times k}$ and $\mathbf{Q}^{(\ell)} \in \mathbb{R}^{\vert \cale \vert \times k'}$, respectively.
Unless otherwise stated, we assume $\mathbf{P}^{(0)} = \mathbf{X}$ and $\mathbf{Q}^{(0)} = \mathbf{Y}$.
We use $\mathbf{I}_{n}$, $\Vert$, $\odot$, and $\sigma(\cdot)$ to denote {the} $n$-by-$n$ identity matrix, vector concatenation, elementwise product, and {a} non-linear activation function, respectively.

%% file: 000_taxonomy_tree.tex
\begin{figure*}[t]{

\centering
    \small
    \resizebox{\textwidth}{!}{
    
\tikzset{
  head/.style  = {draw, text width=5cm, drop shadow, font=\sffamily, rectangle, sibling distance = 1cm, level distance=1cm},
  basic/.style  = {draw, text width=2cm, drop shadow, font=\sffamily, rectangle, sibling distance = 1cm, level distance=1cm},
  root/.style   = {head, rounded corners=2pt, thin, align=center, sibling distance = 1cm, fill=gray!10, level distance=1cm},
  level 21/.style = {basic, rounded corners=6pt, thin,align=center, fill=red!10, text width=3.6cm},
  level 22/.style = {basic, rounded corners=6pt, thin,align=center, fill=orange!10,text width=3.6cm},
  level 23/.style = {basic, rounded corners=6pt, thin,align=center, fill=green!10, text width=3.6cm},
  level 24/.style = {basic, rounded corners=6pt, thin,align=center, fill=cyan!10, text width=3.6cm},
  level 31/.style = {basic, rounded corners=6pt, thin, align=center, fill=red!30, text width=1.4cm},
  level 32/.style = {basic, rounded corners=6pt, thin, align=center, fill=orange!30, text width=1.8cm},
  level 33/.style = {basic, rounded corners=6pt, thin, align=center, fill=green!30, text width=1.6cm, sibling distance = 2.5cm},
  level 34/.style = {basic, rounded corners=6pt, thin, align=center, fill=cyan!30, text width=1.4cm},
  level 41/.style = {basic, rounded corners=6pt, thin, align=center, fill=red!50, text width=1.2cm}, 
  level 42/.style = {basic, rounded corners=6pt, thin, align=center, fill=orange!50, text width=1.6cm},
  level 43/.style = {basic, rounded corners=6pt, thin, align=center, fill=green!50, text width=1.3cm},
  level 44/.style = {basic, rounded corners=6pt, thin, align=center, fill=cyan!50, text width=1.4cm}
}

\begin{tikzpicture}[
  level 1/.style={sibling distance=55mm, level distance=.9cm},
  edge from parent/.style={->,draw},
  >=latex]

\node[root] {Modeling higher-order interactions}
  child {[sibling distance = 1.8cm] node[level 21] (c1) {Encoding: Input feature} 
    child {node[level 31] (c11) {External \\ info.}}
    child {node[level 31] (c12) {Structural \\ info.}}
    child {node[level 31] (c13) {Identity \\ info.}}
  }
  child { [sibling distance = 2.2cm] node[level 22] (c2) {Encoding: Input structure}
    child {node[level 32] (c21) {Reductive \\ transformation}}
    child {node[level 32] (c22) {Non-reductive \\ transformation}}
  }
  child {[sibling distance = 1.83cm] node[level 23] (c3) {Encoding: Message passing}
    child {node[level 33] (c31) {Target \\ selection}}
    child {node[level 33] (c32) {Message \\ representation}}
    child {node[level 33] (c33) {Aggregate \\ function}}
}
  child {[sibling distance = 1.85cm] node[level 24] (c4) {Training: Objective}
    child {node[level 34] (c41) {Learning to \\ classify}}
    child {node[level 34] (c42) {Learning to \\ contrast}}
    child {node[level 34] (c43) {Learning to \\ generate}}
  };

\begin{scope}[every node/.style={level 41}]
\node [below of = c11, xshift=5pt] (c111) {Feature};
\node [below of = c111] (c112) {Label};
\end{scope}

\begin{scope}[every node/.style={level 41}]
\node [below of = c12, xshift=5pt] (c121) {Local};
\node [below of = c121] (c122) {Global};
\end{scope}

\begin{scope}[every node/.style={level 41}]
\node [below of = c13, xshift=5pt] (c131) {Random \\ indicator};
\end{scope}

\begin{scope}[every node/.style={level 42}]
\node [below of = c21, xshift=5pt] (c211) {Clique};
\node [below of = c211] (c212) {Adaptive};
\end{scope}

\begin{scope}[every node/.style={level 42}]
\node [below of = c22, xshift=5pt] (c221) {Star};
\node [below of = c221] (c222) {Line};
\node [below of = c222] (c223) {Tensor};
\end{scope}

\begin{scope}[every node/.style={level 43}]
\node [below of = c31, xshift=3pt] (c311) {Node to \\ node};
\node [below of = c311] (c312) {Node to \\hyperedge};
\end{scope}

\begin{scope}[every node/.style={level 43}]
\node [below of = c32, xshift=3pt] (c321) {Hyperedge \\ consistent};
\node [below of = c321] (c322) {Hyperedge \\ dependent};
\end{scope}

\begin{scope}[every node/.style={level 43}]
\node [below of = c33, xshift=3pt] (c331) {Fixed \\ pooling};
\node [below of = c331] (c332) {Learnable \\ pooling};
\end{scope}

\begin{scope}[every node/.style={level 44}]
\node [below of = c41, xshift=5pt] (c411) {Rule-based \\ neg. sam.};
\node [below of = c411] (c412) {Learnable \\ neg. sam.};
\end{scope}

\begin{scope}[every node/.style={level 44}]
\node [below of = c42, xshift=5pt] (c421) {Node \\ level};
\node [below of = c421] (c422) {Hyperedge \\ level};
\node [below of = c422] (c423) {Membership \\ level};
\end{scope}

\begin{scope}[every node/.style={level 44}]
\node [below of = c43, xshift=5pt] (c431) {Ground \\ truth};
\node [below of = c431] (c432) {Latent} ; 
\end{scope}

\foreach \value in {1,2}
  \draw[->] (c11.195) |- (c11\value.west);

\foreach \value in {1, 2}
  \draw[->] (c12.195) |- (c12\value.west);

\foreach \value in {1}
  \draw[->] (c13.195) |- (c13\value.west);

\foreach \value in {1, 2}
  \draw[->] (c21.195) |- (c21\value.west);

\foreach \value in {1, 2, 3}
  \draw[->] (c22.195) |- (c22\value.west);

\foreach \value in {1,2}
  \draw[->] (c31.195) |- (c31\value.west);

\foreach \value in {1,2}
  \draw[->] (c32.195) |- (c32\value.west);

\foreach \value in {1,2}
  \draw[->] (c33.195) |- (c33\value.west);

\foreach \value in {1,2}
  \draw[->] (c41.195) |- (c41\value.west);

\foreach \value in {1,2, 3}
  \draw[->] (c42.195) |- (c42\value.west);

\foreach \value in {1,2}
  \draw[->] (c43.195) |- (c43\value.west);

  \end{tikzpicture}
  }
  \caption{Taxonomy on modeling higher-order interactions. The term neg. sam. denotes {negative sampling.} 
  \label{fig:overall_taxonomy}}
  }
\end{figure*}
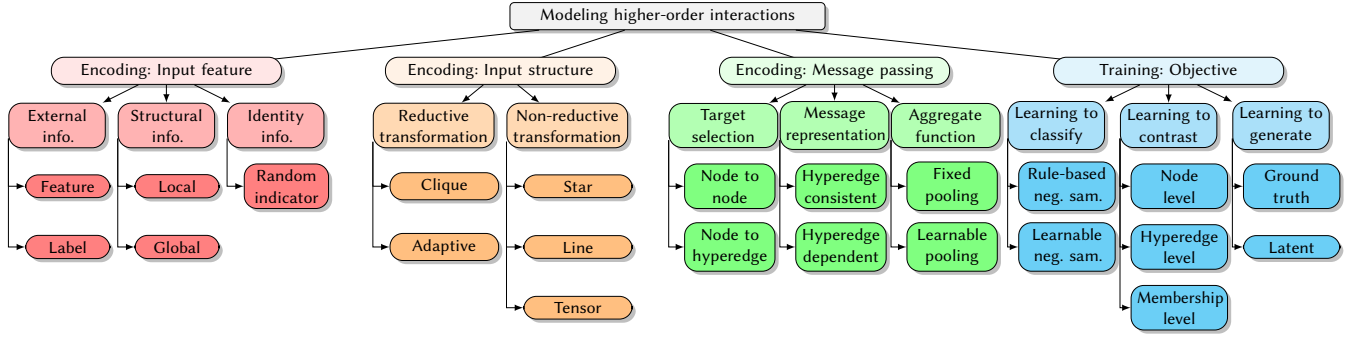

%% file: 000_table_notations.tex
\begin{table}[t!]
\caption{Frequently-used symbols} \label{tab:notations}
    \small
    \centering
    \scalebox{0.75}{
    \renewcommand{\arraystretch}{0.8}
        \centering
        \begin{tabular}{l | l }
            \toprule
            \textbf{Notation} & \textbf{Definition} \\
            \midrule
            \midrule
            $\mathcal{G} = (\mathcal{V}, \mathcal{E})$ & Hypergraph with nodes set $\mathcal{V}$ and hyperedges set $\mathcal{E}$\\
            
            $\mh \in \{0,1\}^{\vert \mathcal{V}\vert \times \vert \mathcal{E} \vert}$ & Incidence matrix\\
            
            $\mx \in \mathbb{R}^{\vert \mathcal{V}\vert \times d}$, $\my \in \mathbb{R}^{\vert \mathcal{E}\vert \times d'}$& Node features ($\mathbf{X}$) and hyperedge features ($\mathbf{Y}$) \\

            $\matp ^{(\ell)} \in \mathbb{R}^{\vert \mathcal{V}\vert \times k}$, $\mq^{(\ell)} \in \mathbb{R}^{\vert \mathcal{E}\vert \times k'}$& $\ell$-th layer embeddings of nodes ($\mathbf{P}^{(\ell)}$) and hyperedges ($\mq^{(\ell)}$) \\

            $\mathcal{N}_{\mathcal{E}}(v_{i})$ & Incident hyperedges of node $v_{i}$\\

            $\mathbf{I}_{n}$ & $n$-by-$n$ identity matrix\\

            $\mathbb{I}[\texttt{cond}]$ & Indicator function that returns 1 if \texttt{cond} is \texttt{True}, 0 otherwise \\



            $\sigma(\cdot)$ & Non-linear activation function \\



            $\mathbf{M}_{i, :} \coloneqq \vm_{i}$ & $i$-th row of matrix $\mathbf{M}$\\  

            $\mathbf{M}_{i,j} \coloneqq m_{ij}$ & $(i,j)$-entry of matrix $\mathbf{M}$\\  
            
            \bottomrule
        \end{tabular}
        }
\end{table}

%% file: 003encoding.tex
In this section, we provide a step-by-step description of how HNNs encode higher-order interactions (HOIs).

\subsection{Step 1: Design features to reflect HOIs}\label{subsec:features}

First, HNNs require a careful choice of input node features $\mathbf{X} \in \mathbb{R}^{\vert \calv \vert \times d}$ and/or hyperedge features $\mathbf{Y} \in \mathbb{R}^{\vert \cale \vert \times d'}$. 
Their quality can be vital for a successful application of HNNs~\cite{geon2024VilLain, zhang2020hyper}.
Thus, studies have crafted input features to enhance HNNs in encoding HOIs. 
Three primary approaches include the use of (\textit{i}) external features or labels, (\textit{ii}) structural features, and (\textit{iii}) identity features.

\subsubsection{\textbf{External features or labels}}\label{subsec:externalfeatures}
External features or labels broadly refer to information that is not directly obtained from the hypergraph structure.
Using external features allows HNNs to capture information that may not be transparent in hypergraph structure alone.
When available, using external node features $\mathbf{X}$ and hyperedge features $\mathbf{Y}$ as HNN input is the standard practice.

Some examples of node features from widely-used benchmark datasets are bag-of-words vectors~\cite{yadati2019hypergcn}, TF-IDFs~\cite{dua2017uci}, visual object embeddings~\cite{feng2019hypergraph}, or noised label vectors~\cite{chien2021you}.
Interestingly, as in label propagation, HyperND~\cite{prokopchik2022nonlinear} constructs input node features $\mathbf{X}$ by concatenating external node features with label vectors. 
Specifically, one-hot-encoded label vectors and zero vectors are concatenated for nodes with known and unknown labels, respectively.
Since external hyperedge features are typically missing in the benchmark datasets,
in practice, input features of $e_{j}$ can be obtained by averaging its constituent nodes (i.e., $\vy_{j} = \sum_{v_{k} \in e_{j}} \boldsymbol{x}_{k}/\vert e_{j}\vert$)~\cite{yan2023hypergraph}.

\subsubsection{\textbf{Structural features}}\label{subsec:structuralfeatures}
On top of external features, studies have also utilized structural features as HNN input features.
{Structural features are typically derived from the input hypergraph structure $\mathcal{E}$ to capture structural proximity or similarity between nodes.}
While leveraging them in addition to the structure $\cale$ may seem redundant,
several studies have highlighted their theoretical and empirical advantages, particularly for hyperedge prediction~\cite{wan2021principled} and for {transformer-based HNNs}~\cite{choe2023classification, saifuddin2023topology, liu2023hypergraph}. 

Broadly speaking, studies have leveraged either local or global structural features.
{To capture local structures around each node}, some HNNs use the incidence matrix $\mathbf{H}$ as part of the input features~\cite{zhang2020hyper, wan2021principled, liu2023hypergraph}.
Notably, HyperGT~\cite{liu2023hypergraph} parameterizes its structural node features $\mathbf{X}' \in \mathbb{R}^{\vert \calv \vert \times k}$ and hyperedge features $\mathbf{Y}' \in \mathbb{R}^{\vert \cale \vert \times k}$ as follows: $\mathbf{X}' = \mathbf{H}\mathbf{\Theta}$ and $\mathbf{Y}' = \mathbf{H}^{T}\mathbf{\Phi}$, where $\mathbf{\Theta} \in \mathbb{R}^{\vert \cale \vert \times k}$ and $\mathbf{\Phi} \in \mathbb{R}^{\vert \calv \vert \times k}$ are learnable {weight} matrices.
Some HNNs leverage structural patterns within each hyperedge.
Intuitively, the importance or role of each node may vary depending on hyperedges.
{For instance,} 
WHATsNet~\cite{choe2023classification} uses within-order positional encoding, where node centrality order within each hyperedge serves as edge-dependent node features (detailed in Sec.~\ref{subsec:aggregation}). 
{Also, a study~\cite{moon2023four} utilizes the occurrence of each hypergraphlet (i.e., a predefined pattern of local structures describing the overlaps of hyperedges within a few hops) around each node or hyperedge as input features.}
{Global features based on roles and proximity in the entire hypergraph context have also been adopted.} 
For example, Hyper-SAGNN~\cite{zhang2020hyper} uses a Hyper2Vec~\cite{huang2019hyper2vec} variant to incorporate structural features preserving node proximity. 
VilLain~\cite{geon2024VilLain} leverages potential node label distributions inferred from the hypergraph structure. 
{HyperFeat~\cite{Do2024HGAlign} aims to capture the structural identity of nodes through random walks.}
THTN~\cite{saifuddin2023topology} integrates learnable node centrality, uniqueness, and positional encodings.

\subsubsection{\textbf{Identity features}}\label{subsec:identityfeatures}
Some HNNs use identity features, especially for recommendation applications.
Generally, {identity features refer to features uniquely assigned to each node (and hyperedge), enabling HNNs to learn distinct embeddings for each node (and hyperedge)~\cite{zhu2021neural,you2021identity}.}
{Prior studies have typically used randomly generated features or separately learnable ones~\cite{ji2020dual, xia2021self, xia2022hypergraph, xia2022self}.}

\subsubsection{\textbf{Comparison with GNNs}}
{Graph neural networks (GNNs)} also require node and/or edge features {for representation learning on pairwise graphs}~\citep{yang2016revisiting, gong2019exploiting, yoo2022accurate}, while typical structural features for GNNs~\cite{grover2016node2vec, dwivedi2022graph, wang2022equivariant2} do not focus on HOIs.


\subsection{Step 2: Express hypergraphs to reflect HOIs}\label{subsec:express}
\begin{figure}[t]
    \centering
    \small
    \subfigure[Hypergraph]
    {\includegraphics[width=0.15\textwidth]{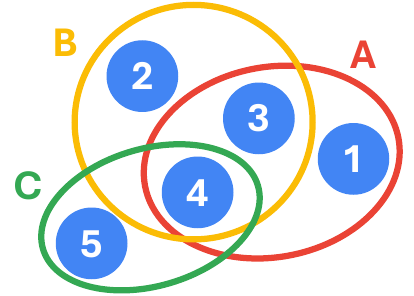}}
    \hspace{1mm}
    \subfigure[Clique-expanded graph with edge weights] 
    {\includegraphics[width=0.15\textwidth]{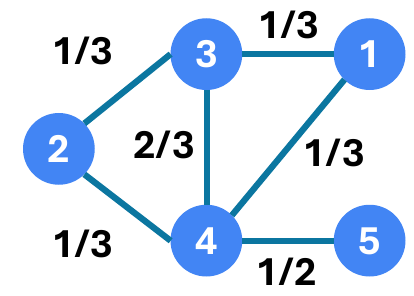}}
    \hspace{1mm}
    \subfigure[Star-expanded graph]
    {\includegraphics[width=0.15\textwidth]{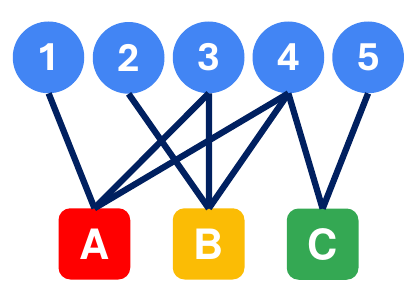}}
    \vspace{-2mm}
    \caption{\label{fig:expression} {An example hypergraph (a), its clique-expanded graph (b), and its star-expanded graph (c).}}
\end{figure}

Some HNNs transform the input hypergraph structure {to better capture the underlying HOIs.} 
They utilize either (\textit{i}) reductive or (\textit{ii}) non-reductive expressions of hypergraph structures (See Fig.~\ref{fig:expression}).

\subsubsection{\textbf{Reductive transformation}}\label{subsec:clique_expansion}
One way to represent hypergraph structure is through \textit{reductive transformation}.
In this approach, each node from the original hypergraph is preserved as a node in the graph, 
while hyperedges are transformed into pairwise edges (see Fig.~\ref{fig:expression}(b)).
Reductive transformation enables the direct application of methods developed for graphs, such as spectral filters~\cite{feng2019hypergraph}, to hypergraphs. However, it may result in {information loss}, and the original hypergraph structure may not be precisely recovered after transformation.
Reductive transformation includes two approaches: clique and adaptive expansion.
Each expansion is represented as $\tau : (\mathcal{E}, \mathbf{X}, \mathbf{Y}) \mapsto \mathbf{A}$, where $\mathbf{A} \in \mathbb{R}^{\vert \mathcal{V}\vert \times \vert \mathcal{V}\vert}$.
We elaborate on the definition of each entry $a_{ij}$ of $\mathbf{A}$ for both expansions.

\smallsection{{Clique expansion.}}
{Clique expansion converts each hyperedge $e_{j} \in \cale$ into a clique (i.e., complete subgraph) formed by the set $e_{j}$ of nodes (see Fig.~\ref{fig:expression}(b)).}
Consider two distinct hypergraphs: ($e_{1} = \{v_{1}, v_{2}, v_{3}\}$) and ($e_{1} = \{v_{1}, v_{2}, v_{3}\}$, $e_{2} = \{v_{1}, v_{3}\}$, and $e_{3} = \{v_{2}, v_{3}\}$).
Despite their changes,
both result in identical clique-expanded graph {($e_{1}=\{v_{1}, v_{2}\}$, $e_{2}=\{v_{1}, v_{3}\}$, and $e_{3}=\{v_{2}, v_{3}\}$)} if edges are unweighted. 
This example illustrates that, in clique expansion, assigning proper edge weights is crucial for capturing HOIs.
To weigh the edges, studies~\cite{feng2019hypergraph, tang2024hypergraph} have utilized (\textit{i}) the node pair co-occurrence, such that pairs appearing together more frequently in hyperedges are assigned larger weights, or (\textit{ii}) hyperedge sizes, such that that node pairs in larger hyperedges are assigned smaller weights.
An example \citep{tang2024hypergraph} is $a_{ij} = \sum\nolimits_{e_{k} \in \mathcal{E}}\frac{\delta(v_{i}, v_{j}, e_{k})}{\vert e_{k}\vert},$
where $\delta(v_{i}, v_{j},e_{k}) = \mathbb{I}[(\{v_{i},v_{j}\} \subseteq e_{k}) \wedge (i \neq j)]$, and $\mathbb{I}[\texttt{cond}]$ is an indicator function that returns $1$ if $\texttt{cond}$ is \texttt{True} and 0 otherwise.

\smallsection{{Adaptive expansion.}}
{Within each transformed clique, some edges may be redundant or even unhelpful. Adaptive expansion} selectively adds {and/or weighs} edges within each clique, often tailored to a given downstream task~\cite{qian2023adaptive, yadati2019hypergcn}.
For example, AdE~\cite{qian2023adaptive} uses a feature-distance-based {edge} weighting strategy.
It obtains projected node features $\mathbf{X}' \in \mathbb{R}^{\vert \calv \vert \times d}$ by $\mathbf{X}' = \mathbf{X} \odot \mathbf{W}$, 
{where all row vectors of $\mathbf{W}$ are $\texttt{sigmoid}(\texttt{MLP}(\sum_{v_{k} \in \calv} \vx_{k}/\vert \mathcal{V}\vert))$.}
Then, AdE selects {two distant nodes $v_{i,j}$ and $v_{k,j}$} within each hyperedge $e_{j}$, i.e., $\{v_{i,j}, v_{k, j}\} = \arg\max_{\{v_{i},v_{k}\} \in \binom{e_{j}}{2}} \vert \sum^{d}_{t=1}(\mathbf{X}'_{i,t} - \mathbf{X}'_{k,t})\vert$.
After that, {it connects the all nodes in $e_{j}$ with $v_{i,j}$ and $v_{k,j}$, essentially adding} $\cale'_{j}= \{\{v_{i,j}, v_{t}\} : v_{t} \in e_{j} \setminus \{v_{i,j}\} \} \cup \{\{v_{k,j}, v_{t}\} : v_{t} \in e_{j} \setminus \{v_{k,j}\} \}$. 
AdE assigns weights to each edge in $\cale'_{j}$ as follows:
\begin{gather}
    a_{ik} = \sum_{e_{j} \in \cale}\frac{\mathbb{I}[\{v_{i},v_{k}\} \in \cale'_{j}]\xi(i,k)}{\sum_{\{v_{s},v_{t}\} \in \binom{e_{j}}{2}}\xi(s,t)},
\end{gather}
where $\xi(i,k) = \exp\left(\lVert \vx_{i} - \vx_{k}\rVert_{2}\sum_{t=1}^{d}{(\mathbf{X}'_{i,t} - \mathbf{X}'_{k,t})^{2}}/{\theta^{2}_{t}}\right)$ and $\theta_{t}$, $\forall t\in [d]$, are learnable scalars.

\subsubsection{\textbf{Non-reductive transformation}}\label{subsec:star_expansion}
Non-reductive transformation of hypergraph structure includes star expansion~\cite{chien2021you, choe2023classification, wang2022equivariant, saifuddin2023hygnn}, line expansion~\cite{yang2022semi}, and tensor representation~\cite{kim2022equivariant, wang2024t, wang2024tensorized}.
They express hypergraph structure without information loss. That is, the hyperedges $\cale$ can be exactly recovered after transformation.

\smallsection{{Star expansion.}}
A \textit{star-expanded graph} of a hypergraph $\calg = (\calv, \cale)$ has two new groups of nodes: the \textit{node group}, which is the same as the node set $\calv$ of $\calg$, and the \textit{hyperedge group}, consisting of nodes corresponding to the hyperedges $\cale$ (refer to Fig.~\ref{fig:expression}(c)).
Star expansion captures HOIs by connecting each node (i.e., a node from the node group) with the hyperedges (i.e., nodes from the hyperedge group) it belongs to, resulting a bipartite graph between the two groups.
Star expansion is expressed as 
$\tau : (\cale, \mathbf{X}, \mathbf{Y}) \mapsto \mathbf{A}$, where each entry of $\mathbf{A} \in \mathbb{R}^{(\vert \mathcal{V}\vert + \vert \mathcal{E}\vert ) \times (\vert \mathcal{V}\vert + \vert \mathcal{E}\vert)}$ is defined as
\vspace{-2mm}
\begin{equation}
    a_{ij} = 
        \begin{cases}
        \mathbb{I}[v_{i} \in e_{j - \vert \mathcal{V} \vert}], & \text{if } 1 \leq i \leq \vert \calv \vert < j \leq \vert \calv \vert + \vert \cale\vert,\\
        \mathbb{I}[v_{j} \in e_{i - \vert \mathcal{V} \vert}], & \text{if } 1 \leq j \leq \vert \calv \vert < i \leq \vert \calv \vert + \vert \cale\vert,\\
        0, & \text{otherwise.}
        \end{cases}
        \vspace{-1mm}
\end{equation}
Here, we assume WLOG that the corresponding index of $v_{i} \in \mathcal{V}$ in $\mathbf{A}$ is $i$, and the corresponding index of $e_{j} \in \mathcal{E}$ in $\mathbf{A}$ is $\vert \calv \vert + j$.

\smallsection{{Line expansion.}}
In a {line-expanded graph} \citep{yang2022semi} of a hypergraph of $\calg = (\calv, \cale)$, each \textit{pair} of a node and a hyperedge containing it is represented as a distinct node. 
That is, its node set is $\{(v_{i}, e_{j}) : v_{i} \in e_{j}, e_{j} \in \cale\}$.
Edges are established between these nodes to connect {each pair of distinct nodes $(v_{i}, e_{j})$ and $(v_{k}, e_{l})$, where $i = k$ or $j=l$}.

\smallsection{{Tensor expression.}}
{Several recent HNNs represent hypergraphs as tensors~\cite{kim2022equivariant, wang2024tensorized}.
For example, T-HyperGNNs~\cite{wang2024t} expresses a $k-$uniform (i.e., $\vert e_{j} \vert = k, \forall e_{j} \in \cale$) hypergraph $\calg = (\calv, \cale)$ with a $k-$order {tensor} $\mathbf{\mathcal{A}} \in \mathbb{R}^{\vert \calv \vert^{k}}$.
That is, if $k=3$, $\mathbf{\mathcal{A}}_{i,j,k} = 1$ if $\{v_{i}, v_{j}, v_{k}\} \in \cale$, and $\mathbf{\mathcal{A}}_{i,j,k} = 0$ otherwise.}


\subsubsection{\textbf{Comparison with GNNs}}
{GNNs typically use the adjacency matrix~\cite{kipf2016semi, wu2019simplifying}, the personalize PageRank matrix~\cite{gasteiger2018predict, chien2020adaptive}, and the Laplacian matrix~\cite{luan2022revisiting} to represent the graph structure.}

\subsection{Step 3: {Pass messages to reflect} HOIs}\label{subsec:messagepassing}

With input features (Sec.~\ref{subsec:features}) and structure (Sec.~\ref{subsec:express}), HNNs learn node (and hyperedge) embeddings. 
They use \textit{neural message passing functions} for each node (and hyperedge) to aggregate messages, i.e., information, from other nodes (and hyperedges).
Three questions arise:
{(\textit{i}) \textit{whose} messages should be aggregated? 
(\textit{ii}) \textit{what} messages should be aggregated? 
(\textit{iii}) \textit{how} should they be aggregated?}

\subsubsection{\textbf{{Whose messages to aggregate}  (target selection)}}\label{subsec:whom2whom}
{For message passing, we should decide whose message to aggregate, typically based on the structural expression of the input hypergraph (Sec.~\ref{subsec:express}).}
{We provide three representative examples: one clique-expansion-based approach and two star-expansion-based ones.}\footnote{
{Regarding target selection,}
adaptive-expansion-~\cite{qian2023adaptive}, line-expansion-~\cite{yang2022hypergraph} and tensor-representation-based~\cite{wang2024tensorized} are similar to clique-expanded ones ($\mathcal{V} \rightarrow \mathcal{V}$).}

\smallsection{{On clique-expanded graphs ($\mathcal{V} \rightarrow \mathcal{V}$)}}. 
Similar to typical GNNs,
clique-expansion-based HNNs perform message passing between neighboring nodes~\cite{feng2019hypergraph, prokopchik2022nonlinear, yadati2019hypergcn, bai2021hypergraph, tang2024hypergraph, benko2024hypermagnet, hayhoe2023transferable}.
They also often incorporate techniques that are effective in applying GNNs.
This is expected since clique expansion transforms a hypergraph into a homogeneous, pairwise graph.
A notable instance is SHNN~\cite{tang2024hypergraph}, which constructs a propagation matrix $\mathbf{W}$ from $\mathbf{A}$ (Sec.~\ref{subsec:clique_expansion}) using a re-normalization trick~\cite{kipf2016semi} as $\mathbf{W} = \tilde{\mathbf{D}}^{-\frac{1}{2}}\tilde{\mathbf{A}}\tilde{\mathbf{D}}^{-\frac{1}{2}}$, where $\tilde{\mathbf{A}} = \mathbf{A} + \mathbf{I}_{\vert \mathcal{V}\vert}$ and $\tilde{\mathbf{D}}$ is the diagonal degree matrix, i.e., $\tilde{\mathbf{D}}_{i,i} = \sum^{\vert \calv\vert}_{k  = 1} \tilde{\mathbf{A}}_{i,k}$.
Then, node embeddings at each $\ell$-th layer are updated using $\mathbf{W}$ as:
\begin{equation}
    \mathbf{P}^{(\ell)} = \sigma\left(((1 - \alpha_{\ell})\mathbf{W}\mathbf{P}^{(\ell - 1)} + \alpha_{\ell} \mathbf{{P}}^{(0)})((1 - \beta_{\ell})\mathbf{I}_{k} + \beta_{\ell} \mathbf{\Theta}^{(\ell)})\right),\label{eq:clique_expansion_mp}
\end{equation}
where $\alpha_{\ell},\beta_{\ell} \in [0,1]$ are hyperparameters, {$\mathbf{\Theta}^{(\ell)} \in \mathbb{R}^{k \times k}$} is a learnable {weight} matrix, and $\mathbf{{P}}^{(0)} = \texttt{MLP}(\mathbf{X})$. 

\smallsection{{On star-expanded graphs} ($\calv \rightarrow \cale$ and $\cale \rightarrow \calv$).}
{In HNNs based on star expansion, message passing occurs from the node group to the hyperedge group ($\calv \rightarrow \cale$) and vice versa ($\cale \rightarrow \calv$)~\cite{wang2022equivariant, chien2021you, choe2023classification, dong2020hnhn, yan2023hypergraph}, either sequentially or simultaneously.}

First, we illustrate sequential message passing using ED-HNN \cite{wang2022equivariant}.
Its message passing at each $\ell$-th layer for each node $v_{i}\in \calv$ is formalized as follows:
\vspace{-3mm}
\begin{align}
    &\vq^{(\ell)}_{j} = \sum_{v_{k} \in e_{j}} \texttt{MLP}_{1}\left(\boldsymbol{p}^{(\ell - 1)}_{k}\right), \label{eq:he_update}\\ 
    &\boldsymbol{\mathscr{p}}^{(\ell)}_{i} = \sum_{e_{k} \in \mathcal{N}_{\cale}(v_{i})} \texttt{MLP}_{2}\left(\left[\vp^{(\ell - 1)}_{i} \lVert \vq^{(\ell)}_{k}\right]\right), \label{eq:v_update} \\ 
    &\vp^{(\ell)}_{i} = \texttt{MLP}_{3}\left(\left[\vp_{i}^{(\ell - 1)} \lVert \boldsymbol{\mathscr{p}}^{(\ell)}_{i} \lVert \vx_{i} \oplus \vert \mathcal{N}_{\cale}(v_{i}) \vert \right]\right), \label{eq:final_update}
\end{align}
where $\vx \oplus c$ denotes the concatenation of vector $\vx$ and scalar $c$. $\texttt{MLP}_{1}$, $\texttt{MLP}_{2}$, and $\texttt{MLP}_{3}$ are MLPs shared across all layers.
Note that, in Eq.~\eqref{eq:he_update}, hyperedge embeddings are updated by aggregating the embeddings of their constituent nodes.
Subsequently, in Eq.~\eqref{eq:v_update} and Eq.~\eqref{eq:final_update}, node embeddings are updated by aggregating transformed embeddings of incident hyperedges.
{Here, the message passing in each direction (Eq.~\eqref{eq:he_update} and Eq.~\eqref{eq:v_update}) occurs sequentially.}


Second, we present an example of simultaneous message passing with HDS$^{ode}$~\cite{yan2023hypergraph}.
Its message passing at each $\ell$-th layer for node $v_{i}\in\calv$ and hyperedge $e_{j}
\in\cale$ is formalized as follows:
\vspace{-2mm}
\begin{align}
    \boldsymbol{\mathscr{p}}^{(\ell)}_{i} &= \vp_{i}^{(\ell - 1)} + \sigma(\vp^{(\ell - 1)}_{i}\mathbf{\Theta}_{(v)} + \vb_{(v)}), \label{eq:simul_mapping1} \\ 
    \boldsymbol{\mathscr{q}}^{(\ell)}_{j} &= \vq_{j}^{(\ell - 1)} + \sigma(\vq^{(\ell - 1)}_{j}\mathbf{\Theta}_{(e)} + \vb_{(e)}), \label{eq:simul_mapping2} \\ 
    \vp^{(\ell)}_{i} &= (1 - \alpha_{(v)})\boldsymbol{\mathscr{p}}^{(\ell)}_{i} + \frac{\alpha_{(v)}}{\vert \caln_{\cale}(v_{i})\vert} \sum\nolimits_{e_{l} \in \caln_{\cale}(v_{i})} \boldsymbol{\mathscr{q}}^{(\ell)}_{l}, \label{eq:simul_passing1}\\ 
    \vq^{(\ell)}_{j} &= (1 - \alpha_{(e)})\boldsymbol{\mathscr{q}}^{(\ell)}_{i} + \frac{\alpha_{(e)}}{\vert e_{j}\vert} \sum\nolimits_{v_{l} \in e_{j}} \boldsymbol{\mathscr{p}}^{(\ell)}_{l}, \label{eq:simul_passing2}
\end{align}
where $\alpha_{(v)},\alpha_{(e)} \in [0,1]$ are hyperparameters, $\mathbf{\Theta}_{(v)}, \mathbf{\Theta}_{(e)} \in \mathbb{R}^{k \times k}$ are learnable weight matrices, and $\boldsymbol{b}_{(v)}, \boldsymbol{b}_{(e)} \in \mathbb{R}^{k}$ are learnable biases.
After projecting node and hyperedge embeddings (Eq.~\eqref{eq:simul_mapping1} and Eq.~\eqref{eq:simul_mapping2}), each node embedding is updated by aggregating the {projected} embeddings of its incident hyperedge  (Eq.~\eqref{eq:simul_passing1}), and each hyperedge embedding is updated by aggregating the {projected} embeddings of its constituent nodes (Eq.~\eqref{eq:simul_passing2}).
{The message passing in each direction (Eq.~\eqref{eq:simul_passing1} and Eq.~\eqref{eq:simul_passing2}) occurs simultaneously}.

\subsubsection{\textbf{What messages to aggregate (message representation)}}\label{subsec:aggregation}
After choosing message targets, the next step is determining \textit{message representations}.
HNNs typically use embeddings from the previous layer as messages, which we term \textit{hyperedge-consistent messages}~\cite{dong2020hnhn, huang2021unignn}.
In contrast, several recent studies propose adaptive message transformation based on its target, which we refer to as \textit{hyperedge-dependent messages}~\cite{choe2023classification, telyatnikov2023hypergraph, aponte2022hypergraph, zheng2024co}.

\smallsection{{Hyperedge-consistent messages.}}
In this widely-used approach \cite{huang2021unignn, chien2021you, yan2023hypergraph}, embeddings from the previous layer are directly treated as vector messages. 
A notable example is UniGNN~\cite{huang2021unignn}, a family of HNNs that obtain node (and hyperedge) embeddings by aggregating the embeddings from its incident hyperedges (or constituent nodes).
UniGIN, a special case of UniGNN, is formalized as follows:

\begin{equation*}
    \vq^{(\ell)}_{j}= \sum_{v_{l} \in e_{j}} \vp_k^{(\ell - 1)} \ ; \ \vp^{(\ell)}_{i} = \left((1 + \epsilon )\vp_{i}^{(\ell - 1)} + \sum_{e_{l} \in \caln_{\cale}(v_{i})} \vq^{(\ell)}_{l}\right)\mathbf{\Theta}^{(\ell)}, 
\end{equation*}
where $\epsilon \in \mathbb{R}$ and $\mathbf{\Theta}^{(\ell)} \in \mathbb{R}^{k \times k'}$ respectively can either be a learnable or fixed scalar and is a learnable weight matrix.

\smallsection{{Hyperedge-dependent messages.}}
The role or importance of a node may vary across the hyperedges it belongs to~\cite{chitra2019random, choe2023classification}.
Several studies~\cite{choe2023classification, telyatnikov2023hypergraph, aponte2022hypergraph} have devised hyperedge-dependent node messages, enabling a node to send tailored messages to each hyperedge it belongs to.
For example, MultiSetMixer~\cite{telyatnikov2023hypergraph} learns different node messages for each incident hyperedge to aggregate with the following message passing function:
\begin{align}
    \vq^{(\ell)}_{j} &= \frac{1}{\vert e_{j}\vert}\sum_{v_{k} \in e_{j}} \vp^{(\ell - 1)}_{k,j} + \texttt{MLP}^{(\ell)}_{1}\left(\texttt{LN}\left(\frac{1}{\vert e_{j}\vert}\sum_{v_{k} \in e_{j}} \vp^{(\ell - 1)}_{k,j}\right)\right), \\ 
    \vp^{(\ell)}_{i,j} &= \vp^{(\ell - 1)}_{i,j} + \texttt{MLP}^{(\ell)}_{2} \left(\texttt{LN}\left(\vp^{(\ell - 1)}_{i,j}\right)\right) + \vq^{(\ell)}_{j},
\end{align}
where $\vp^{(\ell)}_{i,j}$ is the $\ell$-th layer message of $v_{i}$ that is dependent on $e_{j}$, $\texttt{MLP}^{(\ell)}_{1}$ and $\texttt{MLP}^{(\ell)}_{2}$ are MLPs, and $\texttt{LN}$ is layer normalization~\cite{ba2016layer}.

Alternatively, some HNNs update messages based on {hyperedge-dependent} node features.
WHATsNet~\cite{choe2023classification} introduces within-order positional encoding ($\texttt{wope}$) to adapt node messages for each target.
Within each hyperedge, WHATsNet ranks constituent nodes according to their centralities for positional encoding.
Formally, let $\mathbf{F} \in \mathbb{R}^{\vert \calv \vert \times T}$ be a node centrality matrix, where $T$ and $\mathbf{F}_{i,t}$ respectively denote the number of centrality measures (e.g., node degree) and the $t$-th centrality measure score of node $v_{i}$.
The order of an element $c$ in a set $\calc$ is defined as $\texttt{Order}(c,\calc) = \sum_{c' \in \calc } \mathbb{I}[c' \leq c]$.
Then, $\texttt{wope}$ of a node $v_{i}$ at a hyperedge $e_{j}$ is defined as follows:
\begin{equation}
    \texttt{wope}(v_{i},e_{j}) = {\big\Vert}_{t=1}^{T} \frac{1}{\vert e_{j}\vert}\texttt{Order}
    (\mathbf{F}_{i,t}, \{\mathbf{F}_{i,t} : v_{i} \in e_j\}).
\end{equation}
Finally, {hyperedge-}dependent node messages are defined as follows:
\begin{equation}
    \vp^{(\ell)}_{i,j} = \vp^{(\ell)}_{i} + \texttt{wope}(v_{i},e_{j})\mathbf{\Psi}^{(\ell)},
\end{equation}
where $\mathbf{\Psi}^{(\ell)} \in \mathbb{R}^{T \times k}$ is a learnable projection matrix.\footnote{Similarly, 
each hyperedge $e_{j}$'s message to each node $v_{i}$ at the $\ell$-th layer is defined as
$\vq^{(\ell)}_{j,i} = \vq^{(\ell)}_{j} + \texttt{wope}(v_{i},e_{j})\mathbf{\Psi}^{(\ell)}$. WHATsNet aggregates $\{q^{(\ell)}_{k,i} : e_{k} \in \mathcal{N}_{(\cale )}(v_{i})$\} to obtain $\mathbf{p}^{(\ell)}_{i}$ via set attention proposed by \citet{lee2019set}. 
We omit the detailed message passing function since we focus on describing how dependent messages are obtained.}

\subsubsection{\textbf{How to aggregate messages (aggregation function)}} The last step is to decide how to aggregate the received messages for each node (and hyperedge). 

\smallsection{{Fixed pooling.}}
Many HNNs use fixed pooling functions, including summation~\cite{wang2022equivariant, huang2021unignn} and average~\cite{wang2024dynamic, gao2022hgnn+}.
For example, ED-HNN~\cite{wang2022equivariant} uses summation to aggregate the embeddings of constituent nodes (or incident hyperedges), as described in Eq.~\eqref{eq:he_update} and Eq.~\eqref{eq:v_update}.
{Clique-expansion-based HNNs without adaptive edge weights also fall into} this category~\cite{tang2024hypergraph, qu2023hypergraph}.
For example, SHNN~\cite{tang2024hypergraph} uses a fixed propagation matrix $\mathbf{W}$ (see Eq.~\eqref{eq:clique_expansion_mp}) to {aggregate} node embeddings. Specifically, $\vp^{(\ell)}_{i}=\sum_{v_{k} \in \calv}\mathbf{W}_{i,j}\vp^{(\ell - 1)}_{k}$.

\smallsection{{Learnable pooling.}}
Several recent HNNs enhance their pooling functions through attention mechanisms, allowing for weighting messages during aggregation.
Two prominent styles are \textit{target-agnostic attention}~\cite{chien2021you, chai2024hypergraph} and \textit{target-aware attention}~\cite{choe2023classification, saifuddin2023hygnn}.

{Target-agnostic attention functions consider the relations among messages themselves.}
AllSetTransformer~\cite{chien2021you} is an example. 
Denote the embeddings of the incident hyperedges of $v_{i}$ at each $\ell$-th layer as $\mathcal{S}^{(\ell)}(v_{i}) \coloneqq \{\vq^{(\ell)}_{k} : e_{k} \in \mathcal{N}_{\cale}(v_{i})\}$ and its {matrix} expression as $\mathbf{S}^{(\ell,i)} \in \mathbb{R}^{\vert \mathcal{S}^{(\ell)}(v_{i}) \vert \times k}$.
Then, $\vp_{i}^{(\ell)}$ is derived from $\mathbf{S}^{(\ell,i)}$ as follows:
\begin{gather}
    \texttt{MH}(\boldsymbol{\theta}, \mathbf{S})=\Vert_{t=1}^{h}\left(\omega\left(\theta_{t}\left(\texttt{MLP}^{(\ell)}_{t,1}(\mathbf{S})\right)^{T}\right)\texttt{MLP}^{(\ell)}_{t,2}(\mathbf{S})\right), \label{eq:agnoist_attention}\\ 
    \vp^{(\ell)}_{i} = \texttt{LN}\left(\boldsymbol{\mathscr{p}}^{(\ell)}_{i} + \texttt{MLP}^{(\ell)}_{3}\left(\boldsymbol{\mathscr{p}}^{(\ell)}_{i}\right)\right) ; \boldsymbol{\mathscr{p}}^{(\ell)}_{i} = \texttt{LN}\left(\boldsymbol{\theta} + \texttt{MH}\left(\boldsymbol{\theta}, \mathbf{S}^{(\ell, i)}\right) \right), \nonumber
\end{gather}
where $\texttt{LN}$ is layer normalization~\cite{ba2016layer}, $\omega(\cdot)$ is row-wise softmax, $\boldsymbol{\theta} = \Vert_{t=1}^{T} \theta_{t}$ is a learnable vector, and $\texttt{MLP}_{t,1}$, $\texttt{MLP}_{t,2}$, and $\texttt{MLP}_{3}$ are MLPs.
{Note that Eq.~\eqref{eq:agnoist_attention} is a widely-used multi-head attention operation~\cite{vaswani2017attention}, where $\boldsymbol{\theta}$ serves as  queries, and $\mathbf{S}$ serves as keys and values.
This process is target-agnostic since it considers only the global variables $\boldsymbol{\theta}$ and the embeddings $\mathbf{S}$ of incident hyperedges, without considering the embedding of the target $v_{i}$ itself.
}

In target-aware attention approaches, target information is incorporated to compute attention weights.
HyGNN~\cite{saifuddin2023hygnn} is an example, with the following message passing function:
\begin{align}
    \vp_{i}^{(\ell)} &= \sigma\left(\sum_{e_{k} \in \caln_{\cale}(v_{i})} \frac{\texttt{Att}_{(\calv)}^{(\ell)}(\vq^{(\ell  -1)}_{k}, \vp^{(\ell  -1)}_{i}) \vq^{(\ell - 1)}_{k} \mathbf{\Theta}^{(\ell, 1)} }{\sum_{e_{s} \in \caln_{\cale}(v_{i})} \texttt{Att}_{(\calv)}^{(\ell)}(\vq^{(\ell  -1)}_{s}, \vp^{(\ell  -1)}_{i})} \right), \\ 
    \vq_{j}^{(\ell)} &= \sigma\left(\sum_{v_{k} \in e_{j}} \frac{\texttt{Att}_{(\cale)}^{(\ell)}(\vp^{(\ell)}_{k}, \vq^{(\ell  -1)}_{j}) \vp^{(\ell)}_{k} \mathbf{\Theta}^{(\ell, 2)} }{\sum_{v_{s} \in e_{j}} \texttt{Att}_{(\cale)}^{(\ell)}(\vp^{(\ell)}_{s}, \vq^{(\ell  -1)}_{j})} \right).
\end{align}
Here, $\texttt{Att}^{(\ell)}_{(\calv)}(\vq, \vp) = \sigma(\vq^{T}\boldsymbol{\psi}^{(\ell)}_{1}\times \vp^{T}\boldsymbol{\psi}^{(\ell)}_{2}) \in \mathbb{R}$ and $\texttt{Att}^{(\ell)}_{(\cale)}(\vp, \vq) = \sigma(\vp^{T}\boldsymbol{\psi}^{(\ell)}_{3}\times \vq^{T}\boldsymbol{\psi}^{(\ell)}_{4} )\in \mathbb{R}$ are attention weight functions, where $\{\boldsymbol{\psi}^{(\ell)}_{1}, \boldsymbol{\psi}^{(\ell)}_{2}, \boldsymbol{\psi}^{(\ell)}_{3}, \boldsymbol{\psi}^{(\ell)}_{4}\}$ and $\{\mathbf{\Theta}^{(\ell, 1)}, \mathbf{\Theta}^{(\ell, 2)}\}$ are sets of learnable vectors and matrices, respectively.
Note that the attention weight functions consider messages from both sources and targets.
Target-aware attention has also been incorporated into clique-expansion-based HNNs, with HCHA~\cite{bai2021hypergraph} as a notable example.

\input{000_table_HNNs}

\subsubsection{\textbf{Comparison with GNNs}}
GNNs also use neural message passing to aggregate information from other nodes~\cite{gilmer2017neural, lee2023towards, liang2024sign}.
{However, since GNNs typically perform message passing directly between nodes,
they are not ideal for learning hyperedge (i.e., HOI) representations or hyperedge-dependent node representations.}


%% file: 000_table_HNNs.tex


\begin{table}[t!]
\begin{center}
\caption{Summary of hypergraph neural networks (HNNs).} 
\label{tab:hnntaxonomy}
    \small
    \centering
    \setlength{\tabcolsep}{2.7pt}
    \scalebox{0.75}{
    \renewcommand{\arraystretch}{0.7}
        \begin{tabular}{l | c | c | P{5mm} P{5mm} | P{10mm} P{10mm} | P{7mm} P{7mm} }
            \toprule

            \multirow{2}{*}{\textbf{Name}} & \multirow{2}{*}{\textbf{Year}} & \multirow{2}{*}{\textbf{Venue}} & \multicolumn{2}{c | }{{\centering\begin{tabular}{@{}c@{}}\textbf{(Structure)}  \\ \textbf{Reductive?} \end{tabular}}} & \multicolumn{2}{c| }{{\centering\begin{tabular}{@{}c@{}} \textbf{(Embedding Type)} \\ \textbf{Edge Dependent?} \end{tabular}}} & \multicolumn{2}{c}{{\centering\begin{tabular}{@{}c@{}} \textbf{(Aggregation)} \\ \textbf{Learnable?} \end{tabular}}} \\

            {} & {} & {} & \textbf{Yes} & \textbf{No}  &  \textbf{Yes} & \textbf{No} & \textbf{Yes} & \textbf{No}\\
            
            \midrule
            \midrule

            HGNN~\cite{feng2019hypergraph} & 2019 & AAAI & \yes &  &   & \yes &   &\yes \\
            \midrule

            HyperGCN~\cite{yadati2019hypergcn} & 2019 & NeurIPS & \yes &  &   & \yes & \yes &  \\
            \midrule

            HNHN~\cite{dong2020hnhn} & 2020 & ICML &  & \yes  &   & \yes &   & \yes\\
            \midrule
            
            HCHA~\cite{bai2021hypergraph} & 2019 & Pat. Rec. & \yes &   &   & \yes & \yes &  \\
            \midrule

            UniGNN~\cite{huang2021unignn} & 2021 & IJCAI &  & \yes  &   &\yes & \yes &  \\
            \midrule

            HO Transformer~\cite{kim2021transformers} & 2021 & NeurIPS &  &  \yes &   & \yes &  \yes &  \\
            \midrule

            \midrule
            
            AllSet~\cite{chien2021you} & 2022 & ICLR & & \yes &    & \yes & \yes &  \\
            \midrule

            HyperND~\cite{prokopchik2022nonlinear} & 2022 & ICML & \yes &   &   & \yes & \yes &  \\
            \midrule
            
            H-GNN~\cite{zhang2022hypergraph} & 2022 & ICML & \yes &   &   & \yes &  & \yes \\
            \midrule

            EHNN~\cite{kim2022equivariant} & 2022 & ECCV &  & \yes &  & \yes & \yes &  \\
            \midrule

            LE$_{GCN}$~\cite{yang2022hypergraph} & 2022 & CIKM &  & \yes &  & \yes &  & \yes \\
            \midrule
            
            HERALD~\cite{zhang2022hypergraph} & 2022 & ICASSP & \yes &  &  & \yes & \yes &  \\
            \midrule

            HGNN+~\cite{gao2022hgnn+} & 2022 & TPAMI &  & \yes &   & \yes &   & \yes\\
            \midrule 



            \midrule

            ED-HNN~\cite{wang2022equivariant} & 2023 & ICLR & & \yes &    &\yes  &  & \yes\\
            \midrule
            
            PhenomNN~\cite{wang2023hypergraph} & 2023 & ICML & \yes &  &   & \yes &  \yes &  \\
            \midrule
            
            WHATsNet~\cite{choe2023classification} & 2023 & KDD & & \yes & \yes &  & \yes &  \\
            \midrule
            
            SheafHyperGNN~\cite{duta2023sheaf} & 2023 & NeurIPS & \yes &  &  & \yes  & \yes  &   \\
            \midrule

            MeanPooling~\cite{lee2023m} & 2023 & AAAI &  & \yes &  & \yes &   &  \yes \\
            \midrule

            HENN~\cite{hayhoe2023transferable} & 2023 & LoG & \yes &  &   &\yes   &  & \yes   \\
            \midrule

            HyGNN~\cite{saifuddin2023hygnn} & 2023 & ICDE & & \yes &   & \yes & \yes &  \\
            \midrule

            HGraphormer~\cite{qu2023hypergraph} & 2023 & arXiv & \yes &   &   & \yes & \yes &    \\
            \midrule

            MultiSetMixer~\cite{telyatnikov2023hypergraph} & 2023 & arXiv &  & \yes  &  \yes &  &  & \yes   \\
            
            \midrule

            \midrule


            HJRL~\cite{yan2024hypergraph} & 2024 & AAAI &  & \yes & & \yes & & \yes \\
            \midrule
            
            HDE$^{ode}$~\cite{yan2023hypergraph} & 2024 & ICLR & & \yes &   & \yes &  & \yes \\
            \midrule 

            HyperGT~\cite{liu2023hypergraph} & 2024 & ICASSP & & \yes  &  & \yes & \yes &  \\
            \midrule 

            THNN~\cite{wang2024tensorized} & 2024 & SDM & & \yes &  & \yes &  & \yes \\
            \midrule 

            UniG-Encoder~\cite{zou2024unig} & 2024 &  Pat. Rec. & & \yes  &   &\yes  &  & \yes \\
            \midrule 
            
            SHNN~\cite{tang2024hypergraph} & 2024 & arXiv & \yes &   & & \yes &  & \yes \\
            \midrule

            HyperMagNet~\cite{benko2024hypermagnet} & 2024 & arXiv & \yes &  &  & \yes & \yes &   \\
            \midrule

            CoNHD~\cite{zheng2024co} & 2024 & arXiv & & \yes & \yes & & \yes & \\ 
            \midrule 
            
        \end{tabular}
        }
\end{center}
\end{table}

%% file: 004training.tex


In this section, we outline training objectives for HNNs to capture HOIs effectively, particularly when label supervision is weak or absent.
Below, we review three branches: (\textit{i}) learning to classify, (\textit{ii}) learning to contrast, and (\textit{iii}) learning to generate.

\subsection{Learning to classify}

HNNs can learn HOIs by classifying hyperedges~\cite{yadati2020nhp, hwang2022ahp, zhang2020hyper, wan2021principled, ko2023enhancing} as positive or negative.
A positive hyperedge is a ground-truth, ``true'' hyperedge, 
and a negative hyperedge often refers to a heuristically generated ``fake'' hyperedge, considered unlikely to {exist}.
By learning to classify them, HNNs may capture the distinguishing patterns of the ground-truth HOIs.


\subsubsection{\textbf{Heuristic negative sampling.}}
We discuss popular negative sampling (NS) strategies to obtain negative hyperedges~\cite{patil2020negative}: 
%
\begin{itemize}[leftmargin=*]
\item \textbf{Sized NS}: each negative hyperedge contains $k$ random nodes.
\item \textbf{Motif NS}: each negative hyperedge contains a randomly chosen $k$ adjacent nodes.
\item \textbf{Clique NS}: {each negative hyperedge is generated by replacing a randomly chosen node in a positive hyperedge with another randomly chosen node adjacent to the remaining nodes.}
\end{itemize}
Similarly, many HNNs use rule-based NS for hyperedge classification~\cite{yadati2020nhp, hwang2022ahp, zhang2020hyper, wan2021principled, ko2023enhancing}.
Others leverage domain knowledge to design NS strategies~\cite{chen2023teasing, wang2024granularity}.

\subsubsection{\textbf{Learnable negative sampling.}}
Notably, \citet{hwang2022ahp} show that training HNNs with the aforementioned NS strategies may cause overfitting to negative hyperedges of specific types.
This may be attributable to the vast population of potential negative hyperedges, where the tiny samples may not adequately represent this population.
To mitigate the problem, they employ adversarial training of a generator that samples negative hyperedges.

\subsubsection{\textbf{Comparison with GNNs}}
Link prediction on pairwise graphs is a counterpart of the HOI classification task~\cite{zhu2021neural, zhang2018link}.
However, the space of possible negative {edges} significantly differs between them.
In pairwise graphs, the size of the space is $O(\vert \mathcal{V}\vert^{2})$.
However, in hypergraphs, since a hyperedge can contain an arbitrary number of nodes, the size of the space is $O(2^{\vert \mathcal{V}\vert})$, which makes finding representative ``unlikely'' HOIs, or negative hyperedges, more challenging~\cite{hwang2022ahp}.
Consequently, learning the distinguishing patterns of HOIs by classifying the positive and negative hyperedges may be more challenging.

\subsection{Learning to contrast}

Contrastive learning (CL) aims to maximize agreement between data obtained from different views.
{Intuitively, views refer to different versions of the same data, original or augmented.}
Training neural networks with CL has shown strong capacity in capturing the input data characteristics~\cite{jaiswal2020survey}.
For HNNs, several CL techniques have been devised to learn HOIs~\cite{lee2023m, wei2022augmentations, kim2023datasets}.
Here, we describe three steps of CL for HNNs: (\textit{i}) obtaining views, (\textit{ii}) encoding, and (\textit{iii}) computing contrastive loss.

\subsubsection{\textbf{View creation and encoding.}}
First, we obtain views for contrast. 
This can be achieved by augmenting the input hypergraph, using \textit{rule-based}~\cite{lee2023m, kim2023datasets} or \textit{learnable}~\cite{wei2022augmentations} methods.

\smallsection{Rule-based augmentation.} This approach stochastically corrupts node features and hyperedges.
For nodes, an augmented feature matrix is obtained by either zeroing out certain entries (i.e., feature values) of $\mathbf{X}$~\cite{lee2023m, ko2023enhancing} or adding Gaussian noise to them~\cite{qian2024dual}.
{For hyperedges, augmented hyperedges are obtained by excluding some nodes from hyperedges~\cite{lee2023m} or perturbing hyperedge membership} (e.g., changing $e_{i} = \{v_{1}, v_{2}, v_{3}\}$ to $e'_{i} = \{v_{1}, v_{2}, v_{4}\}$)~\cite{liu2023multi}.

\smallsection{Learnable augmentation.} This approach utilizes a neural network to generate views~\cite{wei2022augmentations}. 
Specifically, HyperGCL~\cite{wei2022augmentations} generates synthetic hyperedges $\cale'$ using HNN-based VAE~\cite{kingma2013auto}.

Once an augmentation strategy $\tau: (\mathbf{X},\cale) \mapsto (\mathbf{X}',\cale')$ is decided, a hypergraph-view pair $(\mathcal{G}^{(1)},\mathcal{G}^{(2)})$ can be obtained in two ways:
\begin{itemize}[leftmargin=*]
    \item $\mathcal{G}^{(1)}$ is the original hypergraph with $(\mx, \cale)$, and $\mathcal{G}^{(2)}$ is an augmented hypergraph with $(\mathbf{X}',\mathcal{E}')$, where $(\mathbf{X}',\mathcal{E}') = \tau (\mathbf{X}, \cale)$~\cite{wei2022augmentations}.
    \item {Both $\mathcal{G}^{(1)}$ and $\mathcal{G}^{(2)}$ are augmented by applying 
    $\tau$ to $(\mx, \cale)$~\cite{lee2023m}. They likely differ due to the stochastic nature of $\tau$.}
\end{itemize}

\smallsection{{Encoding.}}
Then, the message passing on the two views {(sharing the same parameters)} results in two pairs of node and hyperedge embeddings denoted by $(\mathbf{P}',\mathbf{Q}')$ and $(\mathbf{P}'',\mathbf{Q}'')$~\cite{lee2023m, ko2023enhancing}.

\vspace{-2mm}

\subsubsection{\textbf{Contrastive loss.}}
Then, we choose a contrastive loss. 
Below, we present \textit{node}-, \textit{hyperedge}-, and \textit{membership}-level contrastive losses. 
Here, $\tau_{x},\tau_{e}, \tau_{m}\in \mathbb{R}$ are hyperparameters.

\smallsection{Node level.} A node-level contrastive loss is used to (\textit{i}) maximize the similarity between the same node from two different views and (\textit{ii}) minimize the similarity for different nodes~\cite{lee2023m, kim2023datasets, ko2023enhancing, wei2022augmentations, ma2023hypergraph}:
\begin{equation}\label{eq:node_cl}
        \mathcal{L}^{(v)}(\mathbf{P}', \mathbf{P}'')=\frac{-1}{\vert \calv\vert}\sum_{v_{i} \in \calv}\log{\frac{\exp(\texttt{sim}(\vp'_{i},\vp''_{i})/\tau_{v})}{\sum_{v_{k} \in \mathcal{V}}\exp(\texttt{sim}(\vp'_{i},\vp''_{k})/\tau_{v})}},
\end{equation}
where $\texttt{sim}(\vx , \vy)$ is the similarity between $\vx$ and $\vy$ (e.g., cosine similarity).

\smallsection{Hyperedge level.} A
hyperedge-level contrastive loss is implemented in a similar manner~\cite{lee2023m, ko2023enhancing, li2024hypergraph}:

\begin{equation}\label{eq:he_cl}
        \mathcal{L}^{(e)}(\mathbf{Q}', \mathbf{Q}'')=\frac{-1}{\vert \cale\vert}\sum_{e_{j} \in \cale}\log{\frac{\exp(\texttt{sim}(\vq'_{j},\vq''_{j})/\tau_{e})}{\sum_{e_{k} \in \cale}\exp(\texttt{sim}(\vq'_{i},\vq''_{k})/\tau_{e})}}.
\end{equation}

\smallsection{Membership level.}
{A membership-level contrastive loss is used to make the embeddings of incident node-hyperedge pairs distinguishable from those of non-incident pairs} across two views~\cite{lee2023m}:
\begin{align*}\label{eq:mem_cl}
    \mathcal{L}^{(m)}(\mathbf{P}', \mathbf{Q}'')&=\frac{-1}{K}\sum_{e_{j} \in \cale} \sum_{v_{i} \in \calv} \underbrace{\mathbb{1}_{i,j}\log{\frac{\exp(\cald(\vp'_{i},\vq''_{j})/\tau_{m})}{\sum_{v_{k} \in \calv}\exp(\cald(\vp'_{k},\vq''_{j})/\tau_{m})}}}_{\texttt{when }\vq''_{j}~\texttt{is an anchor}} \\
    &-\frac{1}{K}\sum_{e_{j} \in \cale} \sum_{v_{i} \in \calv} \underbrace{\mathbb{1}_{i,j}\log{\frac{\exp(\cald(\vq''_{j}, \vp'_{i})/\tau_{m})}{\sum_{e_{k} \in \cale}\exp(\cald(\vq''_{k}, \vp'_{i})/\tau_{m}),}}}_{\texttt{when }\vp'_{i}~\texttt{is an anchor}}
\end{align*}
where $\mathbb{1}_{s,j} = \mathbb{I}[v_{s} \in v_{j}]$; $\cald(\vx , \vy) \in \mathbb{R}$ is a discriminator {for assigning higher value to incident pairs than non-incident pairs}~\cite{velivckovic2018deep}.

\subsubsection{\textbf{Comparison with GNNs}}
GNNs are also commonly trained with contrastive objectives~\cite{velivckovic2018deep, qiu2020gcc, you2020graph}.
They typically focus on node-level~\cite{velivckovic2018deep} and/or graph-level contrast~\cite{qiu2020gcc}.

\subsection{Learning to generate}

HNNs can also be trained by learning to generate hyperedges. 
Existing HNNs aim to generate either (\textit{i}) ground-truth hyperedges to capture their characteristics or (\textit{ii}) latent hyperedges potentially beneficial for designated downstream tasks.

\subsubsection{\textbf{Generating ground-truth HOIs.}}\label{subsec:gengroundtruthhoi}
Training neural networks to generate input data has shown strong efficacy in various domains and downstream tasks~\cite{openai2023gpt4, he2022masked}. 
In two recent studies, HNNs are trained to generate ground-truth hyperedges to learn HOIs~\cite{kim2023hypeboy, du2022self}.
HypeBoy by~\citet{kim2023hypeboy} formulates hyperedge generation as a \textit{hyperedge filling task}, where the objective is to identify the missing node for a given subset of a hyperedge. 
{Overall}, HypeBoy involves three steps: (\textit{i}) hypergraph augmentation, (\textit{ii}) {node and hyperedge-subset encoding}, and (\textit{iii}) loss-function computation. 

{HypeBoy obtains the augmented node feature matrix $\mathbf{X}'$ and augmented input topology $\cale'$, respectively by randomly masking some entries of $\mathbf{X}$ and by randomly dropping some hyperedges from $\cale$.}
{Hypeboy, then, feeds $\mathbf{X}'$ and $\mathcal{E}'$ into an HNN to obtain node embedding matrix $\mathbf{P}$.}
Subsequently, for each node $v_{i} \in e_{j}$ and subset $q_{ij} = e_{j} \setminus \{v_{i}\}$, HypeBoy obtains (final) node embedding $\boldsymbol{\mathscr{p}}_{i} = \texttt{MLP}_{1}(\vp_{i})$ 
and subset embedding $\boldsymbol{\mathscr{q}}_{ij} = \texttt{MLP}_{2}(\sum_{v_{k} \in q_{ij}} \vp_{k})$.
{Lastly, the HNN is trained to make embeddings of the `true' node-subset pairs similar and of the `false' node-subset pairs dissimilar. 
Specifically, it minimizes the following loss:}
\begin{equation}
    \mathcal{L} = -\sum_{e_{j} \in \cale}\sum_{v_{i} \in e_{j}} \log{\frac{\exp(\texttt{sim}(\boldsymbol{\mathscr{p}}_{i}, \boldsymbol{\mathscr{q}}_{ij}))}{\sum_{v_{k} \in \calv} \exp(\texttt{sim}(\boldsymbol{\mathscr{p}}_{k}, \boldsymbol{\mathscr{q}}_{ij}))}},
\end{equation}
where $\texttt{sim}(\vx, \vy)$ is a cosine similarity between $\vx$ and $\vy$.

\subsubsection{\textbf{Generating latent HOIs.}}\label{subsec:genlatenthoi}
{HNNs can be trained to generate latent hyperedges, especially when} (\textit{i}) {(semi-)}supervised downstream tasks 
and (\textit{ii}) suboptimal input hypergraph structures are assumed.
{Typically, the training methods let HNNs generate potential, latent hyperedges, which are used for message passing to improve downstream task performance~\cite{zhang2022learnable, zhang2022deep, cai2022hypergraph, lei2024unveiling}.}

For example, HSL~\cite{cai2022hypergraph} adopts a learnable augmenter to replace unhelpful hyperedges with the generated ones.
HSL prunes hyperedges using a masking matrix $\mathbf{M} \in \mathbb{R}^{\vert \calv \vert \times \vert \cale \vert}$. 
Each $j-$th column is $m_{j} = \texttt{sigmoid}((\log{(\frac{z_{j}}{1-z_{j}})} + (\epsilon_{0} - \epsilon_{1}))/\tau)$, where $\epsilon_{0}$ and $\epsilon_{1}$, $\tau \in \mathbb{R}$, and $z_{k} \in [0, 1], \forall e_{k} \in \cale$ respectively are random samples from \texttt{Gumbel(0, 1)}, a hyperparameter, and a learnable scalar.
An unhelpful $e_{k}$ is expected to have small $z_{k}$ to be pruned. 

After performing pruning by $\mathbf{\hat{H}} = \mathbf{H} \odot \mathbf{M}$, HSL modifies $\mathbf{\hat{H}}$ by {adding generated latent hyperedges $\mathbf{\Delta H}$.}
Specifically, $\mathbf{\Delta H}_{i,j} = 1$ if $(\mathbf{H}_{i, j } = 0) \wedge (\mathbf{S}_{i,j} \in \texttt{top}(\mathbf{S}, N))$, and 0 otherwise. 
$\texttt{top}(\mathbf{S}, N)$ denotes the set of top-N entries in a learnable score matrix  $\mathbf{S} \in \mathbb{R}^{\vert \calv \vert \times \vert \cale \vert}$.
Each score in $\mathbf{S}$ is $\mathbf{S}_{i, j} = \frac{1}{T}\sum_{t=1}^{T}\texttt{sim}(\vw_{t} \odot \vp_{i} , \vw_{t} \odot \vq_{i})$, where $\{\vw_{t}\}^{T}_{t=1}$ and $\texttt{sim}$ respectively are learnable vectors and cosine similarity. 
{To summarize, node and hyperedge similarities learned by an HNN serve to generate latent hyperedges $\mathbf{\Delta H}$.}
Lastly, $\mathbf{\hat{H}} + \mathbf{\Delta H}$ is fed into {another} HNN for a target downstream task (e.g., node classification). 
All learnable components, {including the HNN for augmentation}, are trained end-to-end.

Note that the HNNs learning to generate latent hyperedges generally implement additional loss functions to encourage the latent hyperedges to be similar to the original ones~\cite{zhang2022learnable,zhang2022deep}.
Furthermore, some studies have explored generating latent HOIs when input hypergraph structures were not available~\cite{jiang2019dynamic, zhou2023totally, gao2020hypergraph}.

\subsubsection{\textbf{Comparison with GNNs}}
Various GNNs also target to generate ground-truth pairwise interactions~\cite{kipf2016variational, tan2023s2gae} or latent pairwise interactions~\cite{fatemi2021slaps}.
In a pairwise graph, the inner product of two node embeddings is widely used to model the likelihood {that an edge joins these nodes}~\cite{kipf2016variational, fatemi2021slaps}.
{However, modeling the likelihood of a hyperedge, which can join any number of nodes, using an inner product is not straightforward.}

%% file: 005applications.tex
HNNs have been adopted in various applications, including recommendation, bioinformatics and medical science, time series analysis, and computer vision.
Their central concerns involve hypergraph construction and hypergraph learning task formulation.

\subsection{Recommendation}

\subsubsection{\textbf{Hypergraph construction.}}
{For recommender system applications, many studies utilized hypergraphs consisting of item nodes (being recommended) and user hyperedges (receiving recommendations).}
For instance, all items that a user interacted with were connected by a hyperedge~\cite{wang2020next}.
When sessions were available, hyperedges connected item nodes by their context window~\cite{wang2021session, xia2021self, li2022enhancing}.
Some studies leveraged multiple hypergraphs. 
For instance, ~\citet{zhang2021double} incorporated user- and group-level hypergraphs. 
\citet{ji2020dual} constructed a hypergraph with item nodes and a hypergraph with user nodes, where their hyperedges were inferred from heuristic-based algorithms.
In contrast, other studies incorporated learnable hypergraph structure~\cite{xia2022hypergraph, xia2022self}.

\subsubsection{\textbf{Application tasks.}}
Hypergraph-based modeling allows natural applications of HNNs for recommendation, typically formulated as a hyperedge prediction problem.
HNNs have been used for 
sequential~\cite{li2021hyperbolic, wang2020next},
session-based \cite{wang2021session, li2022enhancing, xia2021self}, 
group~\cite{zhang2021double, jia2021hypergraph}, 
conversational~\cite{zhao2023multi}, 
and point-of-interest~\cite{lai2023multi} recommendation.

\subsection{Bioinformatics and medical science}

\subsubsection{\textbf{Hypergraph construction.}}
For bioinformatics applications, molecular-level structures have often been regarded as nodes. 
Studies used hyperedges to connect the structures based on their
joint reaction~\cite{chen2023teasing}, 
presence within each drug~\cite{saifuddin2023hygnn, hu2024dual},
and association with each disease~\cite{hu2024dual}.
Some studies used multiple node types. 
A study considered cell line nodes and drug nodes,
with hyperedge connecting those with a synergy relationship~\cite{liu2022multi, wang2024granularity}.
Drugs and their side effects were also considered as nodes, where a hyperedge connected those with drug-drug interaction~\cite{nguyen2022sparse}.
Drug nodes or target protein nodes were also connected by hyperedges based on their similarity in interactions or associations~\cite{ruan2021exploring}.
{Studies also used kNN or learnable hyperedges to build hypergraphs~\cite{peng2024mhclmda, li2023scmhnn}.}

Some other studies used hypergraphs to model MRI data.
Many of them had a region-of-interest serving as a node, while a hyperedge connected the nodes using interaction strength estimation~\cite{wang2023dynamic}, k-means~\cite{ji2022fc}, or random-walk-based sampling~\cite{cai2023discovering}.
On the other hand, in some studies, study subjects were nodes, and hyperedges connected the neighbors found by kNN~\cite{hao2024hypergraph, madine2020diagnosing}.

Lastly, electronic health records (EHR) data were often modeled with hypergraphs.
Most often, nodes were either medical codes~\cite{cai2022hypergraph_medical, wu2023megacare, xu2022counterfactual, xu2023hypergraph, cui2024multimodal} or clinical events~\cite{zhu2022temporal}.
A hyperedge connected the codes or clinical events that were shared by each patient.

\subsubsection{\textbf{Application tasks.}}

For bioinformatics applications, HNNs have been applied to predict interactions or associations among molecular-level structures. 
Thus, many of the tasks could be naturally formulated as a hyperedge prediction task.
Specifically, the application tasks include predictions of 
missing metabolic reactions~\cite{yadati2020nhp,chen2023teasing},
drug-drug interactions~\cite{liu2022multi, wang2024granularity, saifuddin2023hygnn, nguyen2022sparse},
drug-target interactions~\cite{ruan2021exploring},
drug-gene interactions~\cite{tao2023prediction},
herb–disease associations~\cite{hu2024dual},
and miRNA-disease associations~\cite{peng2024mhclmda}.

For MRI analysis, when a region-of-interest served as a node, HNNs have been applied to solve a hypergraph classification problem.
Alzheimer's disease classification~\cite{hao2024hypergraph},
brain connectome analysis~\cite{wang2023dynamic},
autism prediction~\cite{madine2020diagnosing, ji2022fc},
and brain network dysfunction prediction~\cite{cai2023discovering} problems
have been solved with HNNs.

In analyzing EHR data, since a hyperedge consisted of medical codes or clinical events of a patient, 
HNNs have been applied for hyperedge prediction.
Studies used HNNs to 
predict 
mortality~\cite{cai2022hypergraph_medical}, 
readmission~\cite{cai2022hypergraph_medical},
diagnosis~\cite{wu2023megacare},
medication~\cite{wu2023megacare},
phenotype~\cite{xu2022counterfactual, cui2024multimodal}, 
clinical outcomes~\cite{xu2022counterfactual, xu2023hypergraph, cui2024multimodal},
and clinical pathways~\cite{zhu2022temporal}.

\subsection{Time series analysis}

\subsubsection{\textbf{Hypergraph construction.}}
A variety of nodes have been used for time series forecast applications.
Depending on the data, nodes were  
cities~\cite{yi2020hypergraph, wang2024dynamic},
gas regulators~\cite{yi2020hypergraph},
rail segments~\cite{yi2020hypergraph},
train stations~\cite{wang2024dynamic},
stocks~\cite{liao2024stock, sawhney2021stock},
or regions~\cite{li2022spatial}.
Studies often leveraged similarity- or proximity-based hyperedges~\cite{yi2020hypergraph, liao2024stock, sawhney2021stock} or learnable hyperedges~\cite{wang2024dynamic, liao2024stock, li2022spatial}.

\subsubsection{\textbf{Application tasks.}}
When applying HNNs, many time series forecast problems can be formulated as node regression problems.
Specifically, the prior works used HNNs to forecast
taxi demands~\cite{yi2020hypergraph},
gas pressures~\cite{yi2020hypergraph},
vehicle speeds~\cite{yi2020hypergraph},
traffic~\cite{luo2022directed, shang2024mshyper, wang2024dynamic, wu2023not, zhao2023dynamic}, 
electricity consumptions~\cite{shang2024mshyper, wu2023not},
meteorological measures~\cite{shang2024mshyper, wang2024dynamic},
stocks~\cite{liao2024stock, sawhney2021stock},
and crimes~\cite{li2022spatial}.

\subsection{Computer vision}

\subsubsection{\textbf{Hypergraph construction.}}
Hypergraph-based modeling has also been adopted for computer vision applications.
Studies used nodes to represent 
image patches~\cite{han2023vision},
features~\cite{yan2020learning}, 
3D shapes~\cite{bai2021multi},
joints~\cite{zhou2022hypergraph, liu2020semi, xu2022adaptive},
and humans~\cite{huang2023reconstructing}.
To connect the nodes by a hyperedge, 
kNN~\cite{yan2020learning, bai2021multi}, 
Fuzzy C-Means~\cite{han2023vision}, 
and other learnable functions~\cite{wadhwa2021hyperrealistic, liu2020semi, xu2022adaptive} were adopted.

\subsubsection{\textbf{Application tasks.}}
For computer vision tasks,
studies used HNNs to solve problems including
image classification~\cite{han2023vision},
object detection~\cite{han2023vision},
video-based person re-identification~\cite{yan2020learning},
image impainting~\cite{wadhwa2021hyperrealistic},
action recognition~\cite{zhou2022hypergraph},
pose estimation~\cite{liu2020semi, xu2022adaptive},
3D shape retrieval and recogntion~\cite{bai2021multi},
and multi-human mesh recovery~\cite{huang2023reconstructing}.
Due to the heterogeneity of the applied tasks, we found no consistent hypergraph learning task formulation.

%% file: 006discussions.tex
In this work, we provide a survey on hypergraph neural networks (HNNs), with a focus on how they address higher-order interactions (HOIs).
We aim for the present survey to be in-depth, covering HNN encoders (Sec.~\ref{sec:encoding}), training objectives (Sec.~\ref{sec:training}), and applications (Sec.~\ref{sec:application}).
Having reviewed the exponentially growing literature, we close the survey with some future directions.

\smallsection{HNN theory.}
Studies have theoretically investigated graph neural networks (GNNs) on their graph isomorphism recognition~\cite{xu2018powerful, vignac2020building}, approximation abilities~\cite{keriven2019universal, maron2019universality}, and relation to homophily~\cite{Ma2022IsNetworks, lee2024feature}.
However, given the complex nature of hypergraphs, directly applying these theoretical findings to hypergraphs can be non-trivial~\cite{feng2024hypergraph}.
Therefore, many theoretical properties of HNNs remain yet to be unveiled, and some areas have begun to be explored, including their generalization abilities~\cite{luo2023expressiveness} and transferability~\cite{hayhoe2023transferable}.

\smallsection{{Advantages of HNNs.}}
Instead of leveraging HNNs, one could use GNNs for a hypergraph by reducing its structure to a pairwise one. 
While studies have empirically shown that HNNs outperform these alternatives~\cite{feng2019hypergraph,chien2021you, dong2020hnhn, wang2022equivariant, kim2023hypeboy}, the {factors that confer HNNs the advantages remain unclear.}
While the advantages of using HOIs for a heuristic classifier have been investigated~\cite{yoon2020much}, 
studies dedicated to HNNs may inspire improved HNNs and their training strategies.

\smallsection{Complex hypergraphs.}
Networks of HOIs often exhibit temporal, directional, and heterogeneous properties, which are respectively modeled by
temporal~\cite{lee2023temporal}, directed~\cite{gallo1993directed}, and heterogeneous~\cite{huang2024link, yadati2020neural} hypergraphs. 
Although their structural patterns have been studied~\cite{lee2023temporal, kim2023reciprocity, moon2023four, benson2018sequences}, 
developing HNNs to learn such complex HOIs is in the early stages~\cite{tran2020directed, luo2022directed, agarwal2022think, huang2024link, yadati2020neural, zhou2023dynamic}. 
Thus, more benchmark datasets and tasks for complex hypergraphs are necessary.
The proper datasets and tasks will catalyze studies to develop HNNs that better exploit the complex nature of HOIs.

%% file: 100config.tex
\newcommand{\smallsection}[1]{{\noindent {\bf{\underline{\smash{#1.}}}}}}
\newcommand{\smallsectiontwo}[1]{{\bf{\underline{\smash{#1}}}}}
\newcommand\red[1]{\textcolor{red}{#1}}
\newcommand\blue[1]{\textcolor{blue}{#1}}
\newcommand\cyan[1]{\textcolor{cyan}{#1}}
\newcommand\violet[1]{\textcolor{violet}{#1}}
\newcommand\teal[1]{\textcolor{teal}{#1}}
\newcommand\brown[1]{\textcolor{brown}{#1}}
\newcommand\green[1]{\textcolor{green}{#1}}

\newtheorem{obs}{Observation}
\newtheorem{defn}{Definition}
\newtheorem{problem}{Problem}
\makeatletter
\renewcommand*{\@opargbegintheorem}[3]{\trivlist
      \item[\hskip \labelsep{\textit{#1\ #2}}] \textit{(#3)}\ }
\makeatother

\newcommand{\method}{RASP\xspace}
\newcommand{\vfree}{\textbf{S1. Variation-Free}\xspace}
\newcommand{\variation}{\textbf{S2. Variations}\xspace}
\newcommand{\event}{\textbf{S3. Event Count}\xspace}
\newcommand{\mixedeasy}{\textbf{S4. Mixed-Easy}\xspace}
\newcommand{\mixedhard}{\textbf{S5. Mixed-Hard}\xspace}